\documentclass[lettersize,journal]{IEEEtran}
\usepackage{amsmath,amsfonts}
\usepackage{algorithmic}
\usepackage{algorithm}
\usepackage{array}
\usepackage[caption=false,font=normalsize,labelfont=sf,textfont=sf]{subfig}
\usepackage{textcomp}
\usepackage{stfloats}
\usepackage{url}
\usepackage{verbatim}
\usepackage{graphicx}
\usepackage{cite}
\usepackage{nicematrix}
\usepackage{hhline}
\usepackage{tabularx}
\usepackage{multirow}
\usepackage{changepage}
\usepackage{setspace}
\usepackage{makecell}

\usepackage{amsmath}
\usepackage{booktabs}
\usepackage[hidelinks]{hyperref}
\usepackage[numbers]{natbib}
\usepackage{enumitem}

\begin{document}

\title{TrajFusionNet+: Transformer-Based Prediction of Pedestrian Crossing Intention via Fusion of Trajectory Representations and Scene Graphs}

\author{François G. Landry and Moulay A. Akhloufi\\
Perception, Robotics and Intelligent Machines Research Group (PRIME)\\
Department of Computer Science, Université de Moncton, Canada\\
{\tt \{efl7126, moulay.akhloufi\}@umoncton.ca}
}



\maketitle

\begin{abstract}

The pedestrian crossing intention task involves predicting whether pedestrians are likely to cross the road from the point of view of an autonomous vehicle. We introduce TrajFusionNet+, a novel transformer-based model for pedestrian crossing intention prediction. TrajFusionNet+ combines sequential and visual representations of pedestrian trajectory with a graph-based representation of the scene context in order to predict pedestrian crossing intention. The proposed architecture builds upon our previous model, TrajFusionNet, and comprises three branches: a Sequence Attention Module (SAM), which processes a sequential representation of past and predicted pedestrian trajectories; a Visual Attention Module (VAM), which utilizes a visual representation of the pedestrian trajectories by overlaying observed and predicted bounding boxes onto scene images; and a Graph Attention Module (GAM), which extracts pedestrian-centric graphs from segmented scene images and captures the relational dependencies between pedestrians and traffic elements. TrajFusionNet+ achieves improved state-of-the-art performance on the two most widely used pedestrian crossing intention datasets, PIE and JAAD. Furthermore, we introduce a new evaluation protocol in which models are trained jointly on the PIE and JAAD datasets but evaluated separately on each. Under this setting, TrajFusionNet+ demonstrates superior generalization compared to existing approaches\footnote{The source code is publicly available at: \url{https://github.com/fglandry/TrajFusionNetPlus}}.


\end{abstract}

\begin{IEEEkeywords}
pedestrian crossing intention, transformer, computer vision, deep learning, autonomous driving
\end{IEEEkeywords}

\section{Introduction}

Understanding pedestrian behavior is a critical component of ensuring safety in autonomous driving systems. Although progress has been made in pedestrian detection and in recognizing when pedestrians are actively crossing, predicting future actions remains challenging. Pedestrian behavior is highly dynamic, shaped by interactions with other road users as well as the surrounding environmental context \cite{galvao2023pedestrian}. From an ego-vehicle perspective, occlusions from other pedestrians, vehicles, and infrastructure further complicate the task \cite{zhang2023pedestrian}. Despite these difficulties, predicting crossing intention is vital for enabling autonomous vehicles to brake or reroute proactively in urban environments.

Multiple recent models achieve strong results on datasets such as PIE \cite{rasouli_pie_2019} and JAAD \cite{rasouli_1_are_2017}. However, many of these models have been shown to exhibit substantial overfitting in cross-dataset evaluation scenarios when trained on one dataset and evaluated on another \citep{gesnouin2022assessing}. The prevailing practice in the literature has been to report results on multiple datasets, but typically by training and testing on the same dataset. This approach can inflate reported performance and does not reflect real-world deployment scenarios in autonomous vehicles, where models are generally not retrained for each environment.

In this work, we introduce TrajFusionNet+, a novel model for pedestrian crossing intention prediction that achieves state-of-the-art performance while demonstrating improved generalizability when compared to prior models. TrajFusionNet+ builds upon our previous architecture, TrajFusionNet \citep{landry2025trajfusionnet}. The proposed architecture comprises three branches: (1) a Sequence Attention Module (SAM), which processes a sequential representation of past and predicted pedestrian trajectories and vehicle speed; (2) a Visual Attention Module (VAM), which utilizes a visual representation of the pedestrian trajectories by overlaying observed and predicted bounding boxes onto scene images; and (3) a Graph Attention Module (GAM), which extracts pedestrian-centric graphs from segmented scene images and explicitly models spatial relationships between pedestrians and traffic elements. 

The GAM branch is a new addition in TrajFusionNet+ and is based on the TokenGT graph transformer \cite{kim2022pure} to process pedestrian-centric graphs, which we show to outperform a graph convolutional network baseline within our model. Another enhancement over the previous TrajFusionNet model is the modeling of temporal dependencies in the VAM branch through an encoder-only transformer operating on features extracted from the visual scene context at different timesteps. TrajFusionNet+ achieves improved state-of-the-art performance on the two most widely used pedestrian crossing intention datasets, PIE and JAAD. Additionally, we propose a new evaluation protocol for
pedestrian crossing intention prediction, where models are trained jointly on the PIE and JAAD datasets and evaluated separately on each. This protocol shows superior generalization of TrajFusionNet+ compared to existing state-of-the-art methods.

\section{Related Work}

In this section, we briefly review prior work on pedestrian crossing intention prediction; more comprehensive surveys can be found in \cite{LANDRY2024129105, galvao2023pedestrian}. 

Many approaches in the literature address the pedestrian intention problem by modeling sequential features (such as pose keypoints, bounding boxes, and vehicle speed) using RNN- or transformer-based architectures. Incorporating multiple modalities generally improves performance but often increases inference time. These approaches include LSOP-Net \cite{liu2025long}, a multimodal RNN with temporal attention, and PedFormer \cite{rasouli2022pedformer}, which captures dependencies between modalities with early multi-head cross-attention. Vision transformers have also been applied to pedestrian intention prediction by leveraging self-attention over image patches \citep{lorenzo_4_capformer_2021}.

Graph-based methods have also shown strong performance by modeling interactions through scene graphs. Most methods use graph convolutional networks (GCNs) to learn from spatial or spatiotemporal relationships \cite{yang2024real, song2022pedestrian}. Other methods apply GCNs to pose keypoints \cite{cadena2022pedestrian, yang2023dpcian, xie2025gtranspdm}. In PedAST-GCN, Ling et al. \cite{ling2024pedast} employ graph representations that integrate bounding box and vehicle speed modalities alongside pose keypoints into GCN layers, followed by attention layers.

TrajFusionNet+ builds on our earlier TrajFusionNet architecture \cite{landry2025trajfusionnet}, enhancing its two core components: the Sequence Attention Module (SAM) and the Visual Attention Module (VAM). In the original TrajFusionNet, the SAM branch uses an encoder–decoder transformer to predict future trajectories and vehicle speed, which are subsequently fed into an encoder-only transformer. The VAM branch processes scene images overlaid with observed and predicted bounding boxes using Visual Attention Networks (VANs) \cite{guo2023visual}. Outputs from the VAM and SAM branches are passed through projection layers and then combined via dense layers in a late-fusion fashion.

\section{Method}

\subsection{Problem Formulation}

The problem at hand consists of predicting whether each detected pedestrian in a video sequence will initiate a crossing action within a specific future window, based on a set of observation frames. The pedestrian crossing intention problem is formulated as a binary classification task at time \begin{math}t\end{math}, where the crossing action \begin{math}A_i \in \{ 0, 1 \}\end{math} is predicted for pedestrian \begin{math}i\end{math} given previous observations \begin{math}\boldsymbol{M}_i\end{math}:

\begin{equation*}
\mathrm{\boldsymbol{M}}_i = \{ \mathrm{\boldsymbol{m}}^{t-m}_i, \mathrm{\boldsymbol{m}}^{t-m+1}_i,...,\mathrm{\boldsymbol{m}}^{t}_i \}
\end{equation*}

We follow the benchmark parameters proposed by Kotseruba et al. \cite{kotseruba_4_benchmark_2021} and predict whether the pedestrian will cross between \begin{math}1\end{math} and \begin{math}2\end{math} seconds after time \begin{math}t\end{math}. 

\subsection{Input Modalities}

Our model uses three input modalities: a sequence of pedestrian bounding boxes, a sequence of vehicle speeds, and a sequence of scene images. Accordingly, \begin{math}\boldsymbol{M}_i\end{math} is composed of three tensors:
\begin{math}
\mathrm{\boldsymbol{B}}_i = \{ \mathrm{\boldsymbol{b}}^{t-m}_i, \mathrm{\boldsymbol{b}}^{t-m+1}_i,...,\mathrm{\boldsymbol{b}}^{t}_i \}
\end{math}, the bounding box sequence,
\begin{math}
\boldsymbol{V} = \{ v^{t-m}, v^{t-m+1}, ..., v^{t} \}
\end{math}, the vehicle speed sequence, and \begin{math}
\mathrm{\boldsymbol{I}} = \{ \mathrm{\boldsymbol{i}}^{t-m}, \mathrm{\boldsymbol{i}}^{t-m+1},...,\mathrm{\boldsymbol{i}}^{t} \}
\end{math}, the scene video sequence. Each bounding box \begin{math}\mathrm{\boldsymbol{b}}^{t}_i\end{math} consists of four values: the \begin{math}x\end{math} and \begin{math}y\end{math} coordinates of its top-left and bottom-right corners. Vehicle speed is treated differently depending on the dataset used.  For the PIE dataset \citep{rasouli_pie_2019}, we use the raw speed values directly. For the JAAD dataset \citep{rasouli_1_are_2017}, which provides only categorical speed labels, we encode them ordinally as follows: \begin{math}\{0\end{math}: stopped, \begin{math}1\end{math}: decelerating, \begin{math}2\end{math}: moving slow, \begin{math}3\end{math}: moving fast, \begin{math}4\end{math}: accelerating\begin{math}\}\end{math}.

\subsection{Architecture}

\begin{figure*}
    \begin{center}
        \includegraphics[width=0.7\linewidth]{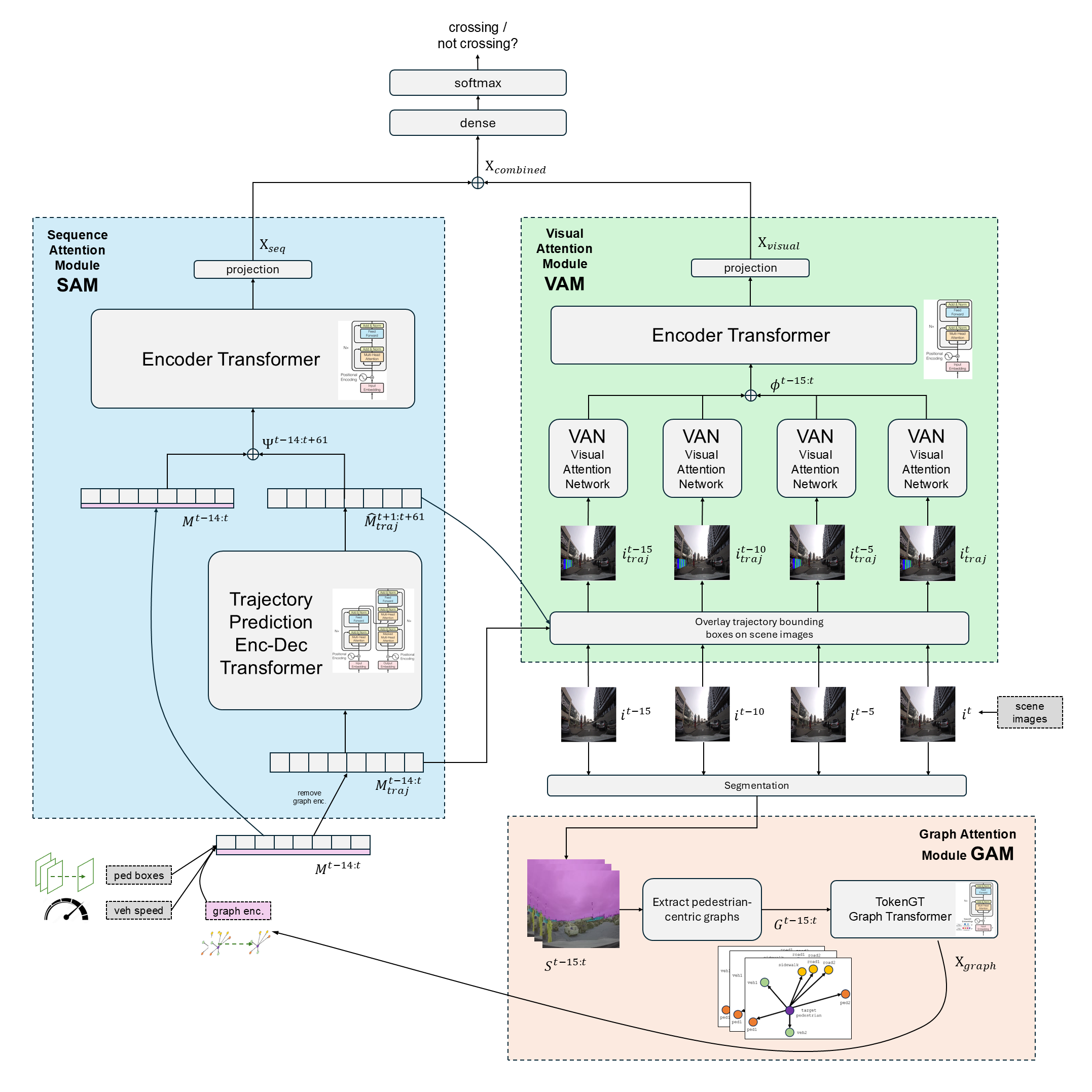}
    \end{center}
   \vspace*{-10pt}
   \caption{TrajFusionNet+ architecture. The model comprises three branches: a Sequence Attention Module (SAM), a Visual Attention Module (VAM), and a Graph Attention Module (GAM). The SAM branch uses an encoder–decoder transformer to predict future pedestrian trajectories and vehicle speed; these predictions are then passed to an encoder-only transformer along with past trajectories and GAM features. The GAM branch extracts pedestrian-centric graph encodings at four timesteps using a TokenGT transformer. The VAM branch applies Visual Attention Networks (VANs) \cite{guo2023visual} to scene images at the same timesteps, overlaid with observed and predicted bounding boxes, followed by temporal modeling using an encoder transformer. Finally, the outputs from the SAM and VAM branches are projected and fused via late fusion.}
\label{fig:figure1_architecture}
\end{figure*}

The proposed architecture is illustrated in Figure~\ref{fig:figure1_architecture}. It consists of three branches: a \textit{Sequence Attention Module (SAM)}, which learns from sequential representations of trajectories; a \textit{Visual Attention Module (VAM)}, which processes video data; and a \textit{Graph Attention Module (GAM)}, which explicitly models spatial relationships between traffic elements through pedestrian-centric graphs. The outputs of these three branches are combined using a late-fusion approach via a dense layer.

\textbf{Sequence Attention Module (SAM)}: The SAM branch closely follows the architecture proposed in the original TrajFusionNet model \citep{landry2025trajfusionnet}. Its role is to extract insights by applying an attention mechanism to the sequential representation of modalities, including past and predicted pedestrian coordinates and vehicle speed, as well as the graph encodings produced by the newly introduced GAM module. The SAM branch consists of two transformer blocks. The first is an encoder-decoder transformer that performs trajectory prediction, estimating future pedestrian bounding boxes and vehicle speed. It is implemented as a non-autoregressive encoder–decoder transformer, as in the original TrajFusionNet model \citep{landry2025trajfusionnet}. It takes as input \begin{math}\boldsymbol{M}_{traj}^{t-14:t}\end{math}, which consists of the past observed pedestrian bounding boxes and vehicle speed, without the graph encodings produced by the GAM branch. The prediction spans the next 60 timesteps, up to 2 seconds after time \begin{math}t\end{math}.

The resulting trajectory prediction tensor, \begin{math}\boldsymbol{\hat{M}}_{traj}^{t+1:t+61} \in \mathbb{R}^{d_{\text{pred}} \times m} \end{math}, does not contain the graph encodings produced by the GAM branch; in order to concatenate it with the past trajectory tensor, \begin{math}\boldsymbol{M}^{t-14:t} \in \mathbb{R}^{d_{\text{seq}} \times (m+g)} \end{math}, we simply append to the former an empty tensor,  \begin{math}\mathbf{0} \in \mathbb{R}^{d_{\text{pred}} \times g} \end{math}. Thus, the input tensor to the transformer encoder, \begin{math}\boldsymbol{\psi}^{t-14:t+61}\end{math}, is obtained as follows:
\begin{equation*}
\boldsymbol{\psi}^{t-14:t+61}
=
\begin{bmatrix}
\boldsymbol{M}^{t-14:t} \\
\big[ \boldsymbol{\hat{M}}_{traj}^{t+1:t+61} \ \ \mathbf{0} \big]
\end{bmatrix}
\end{equation*}

The resulting \begin{math}\boldsymbol{\psi}^{t-14:t+61}\end{math} tensor is fed to the input of an encoder-only transformer whose task is to output an encoding to be used for classification. A final projection layer is added at the end of the SAM branch.

\begin{figure}
    \begin{center}
        \includegraphics[width=1\linewidth]{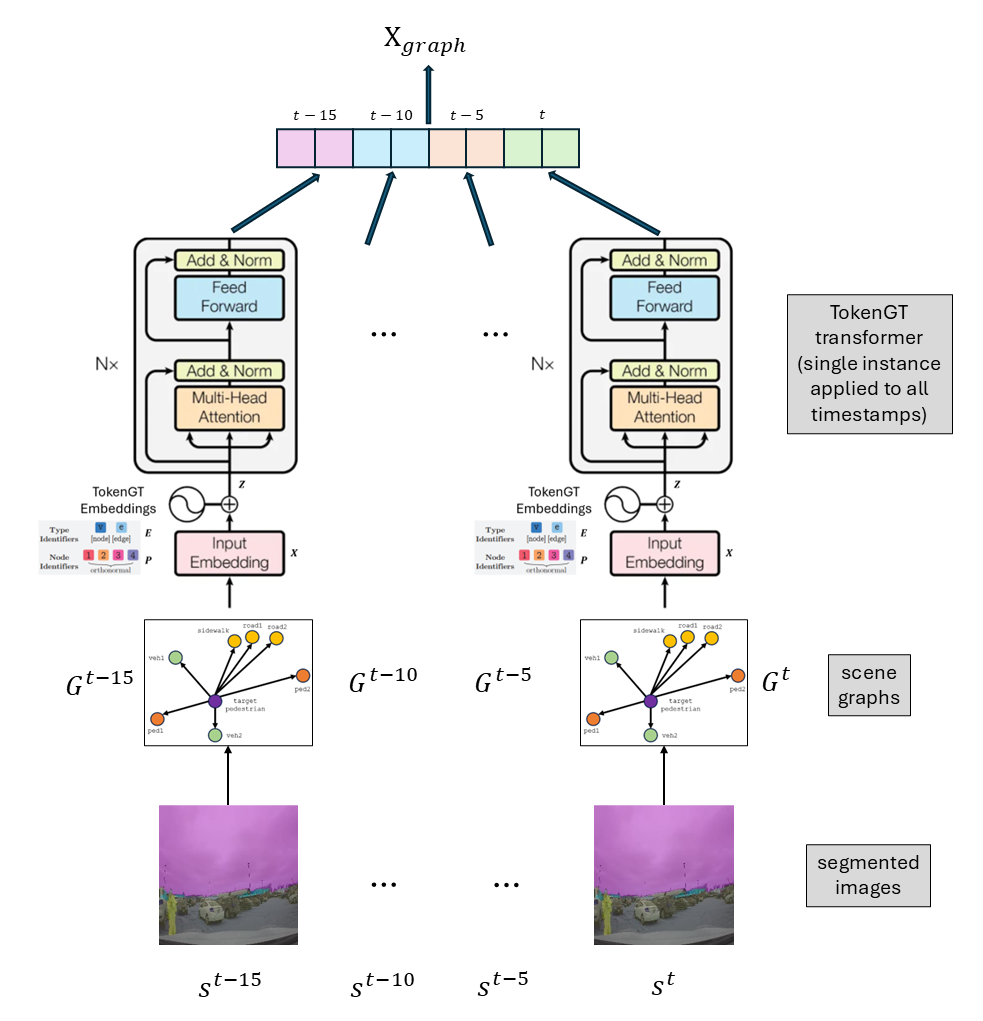}
    \end{center}
   \vspace*{-10pt}
   \caption{Architecture of the Graph Attention Module (GAM)}
\label{fig:figure2_gam_branch}
\end{figure}

\textbf{Graph Attention Module (GAM)}: The role of the GAM branch, illustrated in Figure~\ref{fig:figure2_gam_branch}, is to explicitly model relationships between traffic elements through a pedestrian-centric graph representation, thereby capturing the relational structure of the scene beyond the visual representations learned by the Visual Attention Module (VAM). The GAM branch takes as input a video sequence of ego-vehicle scene images, \begin{math}
\mathrm{\boldsymbol{I}} = \{ \mathrm{\boldsymbol{i}}^{t-m}, \mathrm{\boldsymbol{i}}^{t-m+1},...,\mathrm{\boldsymbol{i}}^{t} \}
\end{math}, extracts pedestrian-centric graphs at selected timesteps, and produces graph encodings using a TokenGT graph transformer \cite{kim2022pure}. These sequential encodings, \begin{math}\boldsymbol{\chi}_{graph} \in \mathbb{R}^{d_{\text{seq}} \times g}\end{math}, are then passed to the encoder transformer in the SAM branch. 

In the first part of the GAM branch, for selected timesteps, we extract a pedestrian-centric undirected graph (following a star topology) in which the central node represents the target pedestrian and the other nodes correspond to traffic elements in the scene. These nodes are obtained from segmentation maps produced by SegFormer \cite{xie2021segformer}. Node features are computed using either the centroid or the closest coordinate (to the target pedestrian) of pixel blobs exceeding a size threshold. Table~\ref{tab:nodes_gam} lists the extracted traffic elements and describes how their features are computed.

\begin{table}[t]
  \centering
  \caption{Nodes Extracted from the Scene Segmentation Map for Pedestrian-Centric Graph Construction}
  \resizebox{\columnwidth}{!}{%
  \begin{tabular}{c cc c}
    \toprule
    Node & Node Name & Node Description & Connected To \\
    \toprule
    \begin{math}n_0\end{math} & \makecell[l]{Target pedestrian} & \makecell{Centroid of pedestrian for which \\we predict crossing or no crossing} & \makecell{\begin{math}n_1\end{math} - \begin{math}n_7\end{math}} \\ \hline 
    \begin{math}n_1\end{math} & Pedestrian 1 & \makecell{Centroid of largest non-target\\ pedestrian detected in the scene} & \begin{math}n_0\end{math} \\ \hline
    \begin{math}n_2\end{math} & Pedestrian 2 & \makecell{Centroid of second-largest non-\\ target pedestrian detected in the scene} & \begin{math}n_0\end{math} \\ \hline
    \begin{math}n_3\end{math} & Road 1 & Centroid of road area & \begin{math}n_0\end{math} \\ \hline
    \begin{math}n_4\end{math} & Road 2 & \makecell{Closest coordinate of road \\area to target pedestrian} & \begin{math}n_0\end{math} \\ \hline
    \begin{math}n_5\end{math} & Sidewalk & \makecell{Closest coordinate of sidewalk \\area to target pedestrian} & \begin{math}n_0\end{math} \\ \hline
    \begin{math}n_6\end{math} & Vehicle 1 & \makecell{Centroid of largest vehicle\\ detected in the scene} & \begin{math}n_0\end{math} \\ \hline
    \begin{math}n_7\end{math} & Vehicle 2 & \makecell{Centroid of second-largest \\vehicle detected in the scene} & \begin{math}n_0\end{math} \\
    \bottomrule    
  \end{tabular}
  }
  \label{tab:nodes_gam}
\end{table}

For each node, the 2D \begin{math}x\end{math} and \begin{math}y\end{math} coordinates are extracted and normalized by the image width and height. For each edge, the distance and angle relative to the target pedestrian are computed, yielding a graph encoding tensor \begin{math}\boldsymbol{G} \in \mathbb{R}^{15 \times 2}\end{math} comprising 8 nodes, 7 edges, and 2 features per node/edge.

To learn from the generated pedestrian-centric graphs, we employ the Tokenized Graph Transformer (TokenGT) proposed by Kim et al. \cite{kim2022pure} instead of traditional graph neural networks. TokenGT is a pure Transformer architecture that represents both nodes and edges as tokens, enabling self-attention to operate directly over the graph structure. TokenGT creates token-wise embeddings composed of \textit{node identifiers} and \textit{type identifiers}. Unlike conventional message-passing graph neural networks, whose information exchange is typically constrained by the graph's connectivity, TokenGT also allows self-attention to model dependencies between arbitrary pairs of nodes and edges. This enables the model to capture potentially relevant interactions that are not explicitly represented in the pedestrian-centric graph, such as secondary interactions between non-target pedestrians.

The \textit{node identifiers} are orthonormal embeddings used to represent the connectivity structure given in the input graph. For a given input graph \begin{math}G = (\mathcal{V}, \mathcal{E})\end{math}, \begin{math}n\end{math} node-wise orthonormal vectors \begin{math}\textbf{p} \in \mathbb{R}^{d_p} \end{math} are produced. Node identifiers are then obtained as follows \cite{kim2022pure}: 


\begin{itemize}[leftmargin=.3in]
  \item For each node \begin{math}v \in \mathcal{V}\end{math}, the node identifier \begin{math}\textbf{P}\end{math} is obtained by concatenating the same orthonormal vector corresponding to the node: \begin{math}\textbf{P} \gets \left[ \textbf{p}_v; \textbf{p}_v \right]\end{math}.
  \item For each edge \begin{math}(u,v) \in \mathcal{E}\end{math}, the node identifier \begin{math}\textbf{P}\end{math} is obtained by concatenating the orthonormal vectors corresponding to each node forming the edge: \begin{math}\textbf{P} \gets \left[ \textbf{p}_u; \textbf{p}_v \right]\end{math}.
\end{itemize}

Kim et al. \cite{kim2022pure} propose to obtain the orthonormal vectors by computing the Laplacian eigenvectors obtained from the eigendecomposition of the graph Laplacian matrix. For the pedestrian-centric graph structure used in TrajFusionNet+, the Laplacian eigenvectors simply become the canonical basis of \begin{math}\mathbb{R}^n\end{math}.

The second component of token-wise embeddings consists of \textit{type identifiers}, which are trainable embeddings that encode whether a token is a node or an edge \cite{kim2022pure}. For a given input graph \begin{math}G = (\mathcal{V}, \mathcal{E})\end{math}, a parameter matrix
\begin{math}\textbf{W}_\textbf{E}=\left[ \textbf{W}_{\textbf{E}^\mathcal{V}}; \textbf{W}_{\textbf{E}^\mathcal{E}} \right]  \in \mathbb{R}^{2 \times d_e}\end{math} is trained, allowing to obtain two embeddings \begin{math} \textbf{E}^\mathcal{V} \end{math} and \begin{math} \textbf{E}^\mathcal{E} \end{math}
for nodes and edges respectively. Type identifiers are obtained as follows:

\begin{itemize}[leftmargin=.3in]
  \item For each node \begin{math}v \in \mathcal{V}\end{math}, the type identifier is \begin{math}\textbf{E} \gets \textbf{E}^\mathcal{V} \end{math}.
  \item For each edge \begin{math}(u,v) \in \mathcal{E}\end{math}, the type identifier is \begin{math}\textbf{E} \gets \textbf{E}^\mathcal{E} \end{math}.
\end{itemize}

Node identifiers \begin{math}\textbf{P}\end{math} and type identifiers \begin{math}\textbf{E}\end{math} are projected into the token dimension \begin{math}d\end{math}. They are then added to token embeddings at the input of the transformer:
\begin{equation*}
\textbf{Z} = \textbf{X} + \textbf{P} + \textbf{E}
\end{equation*}
where \begin{math}\textbf{Z} \in \mathbb{R}^{n \times d} \end{math} is the input sequence to the transformer, \begin{math}\textbf{X} \in \mathbb{R}^{n \times d} \end{math} corresponds to the original token embeddings, \begin{math}\textbf{P} \in \mathbb{R}^{n \times d}\end{math} corresponds to the node identifiers and \begin{math}\textbf{E} \in \mathbb{R}^{n \times d}\end{math} corresponds to the type identifiers. Positional encodings are not used.

In TrajFusionNet+'s GAM branch, TokenGT token-wise embeddings are fed into a standard encoder-only transformer. A single instance of this transformer is shared and applied to the graph encoding tensors (\begin{math}\boldsymbol{G}^t \in \mathbb{R}^{15 \times 2}\end{math}) obtained at selected timesteps. In our implementation, we apply the transformer to four evenly spaced timesteps during the observation period: \begin{math}t-15\end{math}, \begin{math}t-10\end{math}, \begin{math}t-5\end{math}, and \begin{math}t\end{math}. We observed that this reduces the number of segmentation maps generated, saving inference time without significantly affecting performance. To obtain encodings \begin{math}\boldsymbol{\chi}_{graph} \in \mathbb{R}^{d_{\text{seq}} \times g}\end{math} of length \begin{math}d_{\text{seq}}\end{math}, we duplicate the encodings obtained at selected timesteps along the time dimension. The resulting output from the GAM branch, \begin{math}\boldsymbol{\chi}_{graph} \end{math}, is then used as input to the SAM branch.

\textbf{Visual Attention Module (VAM)}: The role of the VAM branch is to extract insights by applying the attention mechanism to visual representations of the pedestrian trajectories as well as the surrounding contextual scene over time. It follows the architecture of the original TrajFusionNet model \citep{landry2025trajfusionnet}, but incorporates an additional encoder-only transformer to capture temporal dependencies between visual encodings. The VAM branch is composed of four Visual Attention Network (VAN) instances \cite{guo2023visual}, applied to scene images at timesteps \begin{math}t-15\end{math}, \begin{math}t-10\end{math}, \begin{math}t-5\end{math}, and \begin{math}t\end{math}. As in the GAM branch, using four timesteps strikes a balance between inference time and predictive performance, as increasing the number of timesteps beyond this yields minimal improvement but increases inference time significantly. The VAN network \cite{guo2023visual} is based on large kernel attention (LKA), which combines the advantages of convolution, such as the ability to capture local structures, with the advantages of self-attention, notably the capacity to model long-range dependencies. In pedestrian crossing intention prediction, local cues such as pedestrian appearance and orientation can be important for understanding behavior, while long-range dependencies may arise from interactions between pedestrians and vehicles that are spatially distant in the ego view.

\begin{figure}
    \begin{center}
        \includegraphics[width=0.5\linewidth]{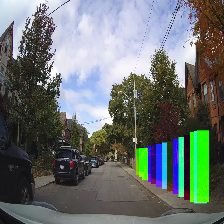}
    \end{center}
   \vspace*{-10pt}
   \caption{Example of scene image (\begin{math}\boldsymbol{i}^{t}_{traj}\end{math}) augmented with rectangles corresponding to the observed (\begin{math}\boldsymbol{M}^{t-14:t}_{1:4}\end{math}) and predicted (\begin{math}\boldsymbol{\hat{M}}^{t+1:t+61}_{1:4}\end{math}) pedestrian bounding boxes.}
\label{fig:figure2_trajectory_van}
\end{figure}

Each VAN processes a scene image augmented with the pedestrian’s observed and predicted bounding boxes, enabling the model to relate the pedestrian’s trajectory to its surrounding context. Bounding boxes are drawn only on two color channels (green and blue in the RGB image) so that the original pedestrian appearance remains available to the network. An example of a scene image augmented with trajectory bounding boxes is shown in Figure~\ref{fig:figure2_trajectory_van}.

The outputs from the four sequential VAN networks are concatenated to form a tensor \begin{math}\boldsymbol{\phi}^{t-15:t} \in \mathbb{R}^{4 \times n}\end{math}, where \begin{math}n=512\end{math} is the dimensionality of each VAN network's output. The tensor \begin{math}\boldsymbol{\phi}^{t-15:t}\end{math} is then passed through an encoder-only transformer to capture temporal dependencies across the VANs' outputs. The transformer's output is subsequently fed into a projection layer, producing the final output of the VAM branch.

\textbf{Late Fusion}: At the end of the TrajFusionNet+ network, the outputs from the Sequence Attention Module (SAM) and the Visual Attention Module (VAM) are merged in a late-fusion fashion using dense layers of sizes 80, 40 and 2.

\subsection{Training and Model Settings}

\textbf{Modular Training}: Given the modularity and relative size of our model, we adopt a staged training strategy in which each lower-level module is pretrained independently. Once trained, the weights of these lower-level modules are frozen, and subsequent modules are fine-tuned on top. During pretraining, each module is attached to a temporary classification head. This head is removed once the module is integrated into the full model.

\textbf{Training Procedure and Learning Objectives}: We first pretrain the trajectory prediction encoder-decoder transformer to predict bounding-box coordinates and vehicle speed for the next 60 timesteps. Training sequences are extracted using the same overlap factor as specified in the benchmark by Kotseruba et al. \cite{kotseruba_4_benchmark_2021}, yielding 49,527 sequences for PIE. The model is trained with a mean squared error (MSE) loss over five z-score normalized values (four bounding-box coordinates and vehicle speed).

All remaining modules are trained as binary classifiers (crossing vs. non-crossing), using the dataset splits and evaluation settings proposed by Kotseruba et al. \cite{kotseruba_4_benchmark_2021}. For PIE, this results in 4,770 training sequences. We use weighted cross-entropy (WCE) as the training objective to address class imbalance.  

We pretrain the TokenGT graph transformer in the GAM branch, shared across selected timesteps, using a classification head consisting of an encoder-only transformer followed by two linear layers. The SAM branch's transformer encoder is then trained with the trajectory predictor and GAM branch frozen. The VAM branch is trained modularly by first training the VAN networks individually, followed by training the VAM transformer encoder on the resulting VAN feature sequences. Finally, we train TrajFusionNet+ end-to-end while freezing all SAM, GAM, and VAM branch weights. Only the projection layers of SAM and VAM and the final dense layers are updated.

\textbf{Implementation and Training Parameters}: For all experiments, we use consistent implementation and training settings across datasets. In the SAM branch's trajectory prediction transformer, we use 8 encoder layers, 8 decoder layers, 4 attention heads, \begin{math}d_{model}=128\end{math}, and 512-dimensional feed-forward layers. All encoder-only transformers in TrajFusionNet+ (SAM, VAM, and TokenGT) use 6 layers, 8 attention heads, \begin{math}d_{model}=128\end{math}, and 1024-dimensional feed-forward layers. The four VAN networks follow the authors' VAN-B2 configuration \cite{guo2023visual}. Training parameters for each module are listed in Table~\ref{tab:training_params}. All modules use linear learning rate decay with warmup; the table reports the peak learning rate. During the final modular training stage, the VAM projection layers' learning rate is set to zero for the first 15 epochs to facilitate more effective learning in the SAM branch.

\begin{table}[t]
  \centering
  \caption{Parameters used for training each module of TrajFusionNet+ during modular training. MSE = mean squared error; WCE = weighted cross-entropy.}
  \resizebox{\columnwidth}{!}{%
  \begin{tabular}{l c ccccc}
    \toprule
    Module & Branch & \makecell{Maximum \\ Learning Rate} & Epochs & Batch Size & Loss \\
    \toprule
    \makecell[l]{Trajectory prediction \\ enc-dec TF} & SAM & 5e-5 & 40 & 64 & MSE \\ \hline 
    Encoder TF & SAM & 5e-6 & 60 & 16 & WCE \\ \hline
    TokenGT graph TF & GAM & 5e-6 & 20 & 16 & WCE \\ \hline
    VAN & VAM & 1e-4 & 15 & 16 & WCE \\ \hline
    Encoder TF & VAM & 5e-6 & 10 & 16 & WCE \\ \hline
    \makecell[l]{Remaining \\ TrajFusionNet+ layers} & --- & 5e-6 & 60 & 8 & WCE \\
    \bottomrule    
  \end{tabular}
  }
  \label{tab:training_params}
\end{table}

\section{Evaluation}

\subsection{Datasets}

We evaluate our model on two benchmark datasets for pedestrian crossing intention prediction: JAAD (Joint Attention for Autonomous Driving) \cite{rasouli_1_are_2017} and PIE (Pedestrian Intention Estimation) \cite{rasouli_pie_2019}. These two datasets are the most popular datasets used in the literature for the prediction of pedestrian crossing intention.

\textbf{JAAD}: A naturalistic driving dataset with 346 video clips of pedestrians prior to crossing. Recorded across North America and Europe under diverse weather conditions, it includes pedestrian bounding boxes and behavioral tags. Vehicle driver behavior is annotated, but numerical vehicle speed is not provided.

\textbf{PIE}: A larger dataset containing nearly ten times more frames, longer pedestrian clips, and richer ego-vehicle information, including speed, GPS, and heading angle. PIE offers more diversity in pedestrian appearance and behavior, but all recordings were in clear weather in Toronto, Canada.

\textbf{Evaluation Procedure}: We adopt the evaluation protocol introduced in the benchmark by Kotseruba et al. \cite{kotseruba_4_benchmark_2021}. We use 0.53 seconds for the observation length (16 frames) and predict pedestrian crossing 1 to 2 seconds (30 to 60 frames) after the observation period. We train and test our model on two datasets: PIE and JAAD{\scriptsize all}, where JAAD{\scriptsize all} refers to the complete JAAD dataset. The results reported for each dataset correspond to the model trained and tested on that specific dataset. We provide results using the following common classification metrics: accuracy (Acc), area under the ROC curve (AUC), F1-score (F1), precision (P), and recall (R).

\subsection{Results}

Table~\ref{tab:comparison_sota} presents a comparison of state-of-the-art (SOTA) models for pedestrian crossing intention prediction. Bold values indicate the best performance across models, while underlined values denote the second-best. TrajFusionNet+ achieves state-of-the-art results, obtaining the highest accuracy and F1-score on the PIE dataset (matching PedFormer \citep{rasouli2022pedformer}), and the second-best scores in both metrics on the JAAD{\scriptsize all} dataset. Compared to its predecessor, TrajFusionNet \citep{landry2025trajfusionnet}, TrajFusionNet+ shows small improvements across several metrics. A key strength of the TrajFusionNet family lies in their consistent performance across datasets and evaluation metrics, unlike models such as PedFormer, which, despite strong results on PIE, obtains a low F1-score on JAAD{\scriptsize all}. This consistency suggests that the TrajFusionNet models are less susceptible to overfitting on specific datasets.

\begin{table*}[t]
  \centering
  \caption{Comparison of state-of-the-art models for the prediction of crossing intention on the PIE and JAAD datasets. Values in bold indicate the best performance, while underlined values represent the second-best performance.}
  \resizebox{0.8\textwidth}{!}{%
  \begin{tabular}{lccccccccccc}
    \toprule
    \multirow{2}{*}{Model} & \multirow{2}{*}{Year} & \multicolumn{5}{c}{PIE} & \multicolumn{5}{c}{JAAD{\scriptsize all}} \\
    \cmidrule(lr){3-7} \cmidrule(lr){8-12}
                                & & Acc & AUC & F1 & P & R & Acc & AUC & F1 & P & R \\ 
    \toprule
    SF-GRU \cite{rasouli_8_pedestrian_2020} & 2020 & 0.82 & 0.79 & 0.69 & 0.67 & 0.70 & 0.84 & 0.84 & 0.65 & 0.54 & \underline{0.84} \\
    PCPA \cite{kotseruba_4_benchmark_2021} & 2020 & 0.86 & \underline{0.91} & 0.78 & 0.69 & \underline{0.89} & 0.83 & 0.77 & 0.57 & 0.50 & 0.66 \\
    TrouSPI-Net \cite{gesnouin2021trouspi} & 2021 & 0.88 & 0.88 & 0.80 & 0.73 & \underline{0.89} & 0.85 & 0.73 & 0.56 & 0.57 & 0.55 \\
    Yang et al. \cite{yang_2_predicting_2022} & 2022 & 0.89 & 0.86 & 0.80 & 0.79 & 0.81 & 0.83 & 0.82 & 0.63 & 0.51 & 0.81 \\
    Pedestrian Graph + \cite{cadena2022pedestrian} & 2022 & 0.89 & 0.90 & 0.81 & 0.83 & 0.79 & 0.86 & \textbf{\textcolor{black}{0.88}} & 0.65 & 0.58 & 0.75 \\
    Song et al. \cite{song2022pedestrian} & 2022 & \underline{0.92} & \underline{0.91} & \underline{0.86} & 0.82 & \textbf{\textcolor{black}{0.90}} & 0.87 & 0.81 & 0.65 & 0.60 & 0.71 \\
    Bai et al. \cite{bai2022deep} & 2022 & 0.89 & 0.88 & 0.79 & 0.74 & 0.84 & 0.86 & 0.81 & \textbf{\textcolor{black}{0.77}} & \textbf{\textcolor{black}{0.74}} & 0.81 \\
    DPCIAN \cite{yang2023dpcian} & 2023 & 0.91 & 0.88 & 0.83 & 0.83 & 0.79 & \underline{0.89} & 0.77 & 0.59 & 0.61 & 0.58 \\
    PIT \cite{zhou_pit_2023} & 2023 & 0.91 & 0.90 & 0.82 & 0.85 & 0.79 & 0.87 & \underline{0.87} & 0.66 & 0.54 & \textbf{\textcolor{black}{0.85}} \\
    PedFormer \cite{rasouli2022pedformer} & 2023 & \textbf{\textcolor{black}{0.93}} & 0.90 & \textbf{\textcolor{black}{0.87}} & \textbf{\textcolor{black}{0.89}} & 0.85 & \textbf{\textcolor{black}{0.93}} & 0.76 & 0.54 & 0.65 & 0.46 \\
    RAIDN \cite{yang2024real} & 2024 & \underline{0.92} & 0.89 & 0.85 & 0.82 & 0.89 & \underline{0.89} & 0.80 & 0.66 & 0.65 & 0.72 \\
    PedAST-GCN \cite{ling2024pedast} & 2024 & 0.91 & \textbf{\textcolor{black}{0.94}} & 0.83 & \underline{0.88} & 0.79 & \underline{0.89} & 0.83 & 0.68 & \underline{0.67} & 0.69 \\
    LSOP-Net \cite{liu2025long} & 2025 & 0.89 & 0.87 & 0.81 & 0.80 & 0.82 & 0.85 & 0.75 & 0.58 & 0.56 & 0.61 \\
    GTransPDM \cite{xie2025gtranspdm} & 2025 & \underline{0.92} & 0.90 & \underline{0.86} & 0.85 & 0.87 & 0.87 & 0.78 & 0.64 & 0.64 & 0.64 \\
    TrajFusionNet \cite{landry2025trajfusionnet} & 2026 & \underline{0.92} & \underline{0.91} & \underline{0.86} & 0.85 & 0.88 & \underline{0.89} & 0.85 & 0.72 & 0.67 & 0.78 \\ 
    \midrule
    TrajFusionNet+ & 2026 & \textbf{\textcolor{black}{0.93}} & \underline{0.91} & \textbf{\textcolor{black}{0.87}} & 0.86 & 0.88 & \underline{0.89} & 0.86 & \underline{0.73} & 0.65 & 0.82 \\ 
    \bottomrule  
  \end{tabular}
  }
  \label{tab:comparison_sota}
\end{table*}

\begin{table*}[t]
  \centering
  \caption{Comparison of state-of-the-art models trained on the combined PIE and JAAD datasets. Models are evaluated on the combined datasets as well as separately on the individual PIE and JAAD datasets. Values in bold indicate the best performance, while underlined values represent the second-best performance.}
  \resizebox{\textwidth}{!}{%
  \begin{tabular}{lccccccccccccccc}
    \toprule
    \multirow{2}{*}{Model} & \multicolumn{5}{c}{Eval on Combined} & \multicolumn{5}{c}{Eval on PIE} & \multicolumn{5}{c}{Eval on JAAD{\scriptsize all}} \\
    \cmidrule(lr){2-6} \cmidrule(lr){7-11} \cmidrule(lr){12-16}
                                & Acc & AUC & F1 & P & R & Acc & AUC & F1 & P & R & Acc & AUC & F1 & P & R \\ 
    \toprule
    C3D \cite{tran2015learning} & 0.71 & 0.68 & 0.49 & 0.41 & 0.62 & 0.65 & 0.62 & 0.47 & 0.41 & 0.56 & 0.77 & 0.75 & 0.52 & 0.41 & 0.72 \\
    SF-GRU \cite{rasouli_8_pedestrian_2020} & 0.73 & 0.77 & 0.59 & 0.45 & \underline{0.85} & 0.78 & \underline{0.80} & \underline{0.69} & 0.57 & \underline{0.86} & 0.67 & 0.74 & 0.47 & 0.33 & 0.84 \\
    PCPA \cite{kotseruba_4_benchmark_2021} & 0.76 & 0.77 & 0.60 & 0.48 & 0.79 & \underline{0.81} & \underline{0.80} & \underline{0.69} & \underline{0.64} & 0.76 & 0.70 & 0.76 & 0.50 & 0.35 & \underline{0.85} \\
    Yang et al. \cite{yang_2_predicting_2022} & 0.41 & 0.60 & 0.43 & 0.28 & \textbf{\textcolor{black}{0.96}} & 0.61 & 0.71 & 0.57 & 0.41 & \textbf{\textcolor{black}{0.93}} & 0.20 & 0.50 & 0.33 & 0.20 & \textbf{\textcolor{black}{1.00}} \\
    TrajFusionNet \cite{landry2025trajfusionnet} & \underline{0.86} & \underline{0.81} & \underline{0.71} & \underline{0.69} & 0.72 & \underline{0.81} & \underline{0.80} & \underline{0.69} & \underline{0.64} & 0.76 & \underline{0.88} & \textbf{\textcolor{black}{0.79}} & \textbf{\textcolor{black}{0.66}} & \underline{0.66} & 0.65 \\ 
    \midrule
    TrajFusionNet+ & \textbf{\textcolor{black}{0.88}} & \textbf{\textcolor{black}{0.82}} & \textbf{\textcolor{black}{0.73}} & \textbf{\textcolor{black}{0.77}} & 0.70 & \textbf{\textcolor{black}{0.88}} & \textbf{\textcolor{black}{0.84}} & \textbf{\textcolor{black}{0.77}} & \textbf{\textcolor{black}{0.81}} & 0.74 & \textbf{\textcolor{black}{0.89}} & \underline{0.78} & \underline{0.65} & \textbf{\textcolor{black}{0.70}} & 0.62 \\ 
    \bottomrule  
  \end{tabular}
  }
  \label{tab:comparison_sota_combined}
\end{table*}

\subsection{Training on Data from Combined Datasets}

To further assess TrajFusionNet+’s generalizability, we propose a protocol in which the model is trained on a combination of the PIE and JAAD{\scriptsize all} datasets and evaluated on each dataset individually. To our knowledge, no prior work in pedestrian crossing intention prediction uses this evaluation setup, although Gesnouin et al. \cite{gesnouin2022assessing} propose a related approach where models are trained on one dataset and evaluated on another. We argue that evaluating models with fixed weights across different datasets provides a more realistic measure of generalizability, as it mirrors deployment scenarios in autonomous vehicles where models are typically not retrained for each environment. Because the JAAD{\scriptsize all} dataset does not provide numerical vehicle speed values, we compare methods without using the vehicle speed as a modality.

Table~\ref{tab:comparison_sota_combined} compares state-of-the-art models using the proposed protocol, with training on combined PIE and JAAD{\scriptsize all} data. Evaluation is performed on the combined datasets as well as separately on the individual PIE and JAAD{\scriptsize all} datasets. Only models with publicly available, runnable code are included.

Overall, evaluation metrics are substantially lower than those in Table~\ref{tab:comparison_sota}, where models were trained and tested separately on each dataset. Although this difference can be partly attributed to the removal of the speed modality, the extent of the performance drop suggests potential overfitting, which was also observed by Gesnouin et al. \cite{gesnouin2022assessing}. Some methods, such as the RNN model by Yang et al. \cite{yang_2_predicting_2022}, perform poorly under this protocol. TrajFusionNet+ achieves the highest performance across most metrics when evaluated on the combined datasets. It outperforms all other models by a large margin on the PIE dataset and performs comparably to the previous TrajFusionNet, but still significantly better than other methods, on the JAAD{\scriptsize all} dataset.

\subsection{Inference Time}
\label{sec:inference_time}

Table~\ref{tab:inference_time} presents a comparison of the inference times of TrajFusionNet+ and several other SOTA approaches. The inference times were measured on a consumer-grade GPU (NVIDIA GeForce RTX 3060). Only methods with publicly available source code were considered. Two types of inference time are reported: model-only (M) and model with data preprocessing included (M + D). Pedestrian detection and tracking are not included in D, as they are shared across all approaches compared.

Data preprocessing includes the time required to compute the input modalities required by each model, such as pose estimation and segmentation maps. For instance, PCPA \cite{kotseruba_4_benchmark_2021} uses OpenPose \cite{cao2017realtime} for pose estimation, which takes 13.99 ms to run on an RTX 3060 GPU. Pedestrian Graph+ \cite{cadena2022pedestrian} and Yang et al. \cite{yang_2_predicting_2022} generate pose keypoints for each observation frame. These two approaches also use DeepLabV3 \cite{chen2017rethinking} to compute segmentation maps (one map for Pedestrian Graph+ and one per observation frame for Yang et al.). DeepLabV3 requires 13.95 ms to execute on the RTX 3060 GPU. 

For TrajFusionNet+, we provide two implementations in Table~\ref{tab:inference_time}: the proposed model, which uses SegFormer \cite{xie2021segformer} as the semantic segmentation backbone in the GAM module, and a lighter-weight implementation that uses DeepLabV3 instead. As shown in Table~\ref{tab:inference_time}, TrajFusionNet+ has a relatively large number of parameters (123.16M) and a model-only inference time (M) that is higher than that of other approaches. However, its total inference time including data preprocessing (M + D) stands at 197.24 ms, which is lower than that of the other approaches considered, except for the original TrajFusionNet. Replacing SegFormer with the lighter DeepLabV3 model further reduces the total inference time (M + D) by decreasing the inference time of each of the four semantic maps generated in the GAM module from 41.16 ms to 13.95 ms on the RTX 3060 GPU.

\begin{table}
  \caption{Inference time obtained by state-of-the-art models and their respective number of parameters. M represents the inference time of the model alone, while M + D represents the inference time of the model combined with data preprocessing.}
  \resizebox{\columnwidth}{!}{%
  \begin{tabular}{lccc}
    \toprule
    \multirow{2}{*}{Model} & \multicolumn{2}{c}{Inference time (ms)} & \multirow{2}{*}{\makecell{Params \\ (millions)}} \\
    \cmidrule(lr){2-3}
                                & M & M + D \\ 
    \toprule
    PCPA \cite{kotseruba_4_benchmark_2021} & 15.51 & 239.35 & 31.17 \\
    Yang et al. \cite{yang_2_predicting_2022} & 2.64 & 449.68 & 2.99 \\
    Pedestrian Graph + \cite{cadena2022pedestrian} & 2.67 & 240.46 & 0.07 \\
    TrajFusionNet \cite{landry2025trajfusionnet} & 12.04 & 12.04 & 58.28 \\
    \midrule
    \makecell[l]{TrajFusionNet+ with\\ DeepLabV3 segmentation} & 32.59 & 88.39 & 123.16 \\
    \midrule
    \makecell[l]{TrajFusionNet+ with\\ SegFormer segmentation\\(proposed model)} & 32.59 & 197.24 & 123.16 \\
    \bottomrule
  \end{tabular}
  }
  \label{tab:inference_time}
\end{table}

\subsection{Ablation Study}

We conducted an ablation study to assess the contribution of individual components within TrajFusionNet+ to the overall model performance. Table~\ref{tab:ablation_study} reports results for scenarios where specific modules were removed or altered.

Starting with the GAM branch, Scenario 1 removes the entire branch, which leads to a small but consistent reduction in performance across most metrics and datasets, showing the branch's importance. Scenario 2 replaces the TokenGT graph transformer with a baseline two-layer GCN, leading to a slight performance drop that shows the advantage of the transformer-based TokenGT for pedestrian-centric graph modeling. Scenario 3 involves replacing SegFormer, the semantic segmentation backbone in the GAM module, with the lighter-weight DeepLabV3 model \cite{chen2017rethinking}, resulting in a small decrease on PIE but little change on JAAD{\scriptsize all}.

For the VAM branch, Scenario 4 simplifies the module to a single VAN network, yielding a minor drop on PIE and a larger one on JAAD{\scriptsize all}, indicating the value of temporal modeling. Scenario 5 removes the overlay of observed and predicted bounding boxes from the scene images, causing a small decrease on PIE and a substantial drop on JAAD{\scriptsize all}, demonstrating the benefit of augmenting the visual input with trajectory information.

For the SAM branch, Scenario 6 removes the predicted trajectory tensor from the transformer input, producing a significant decrease on PIE but minimal change on JAAD{\scriptsize all}. Scenario 7 excludes vehicle speed from the input modalities, leading to a noticeable decrease in performance across metrics on the PIE dataset, and a small decrease on the JAAD{\scriptsize all} dataset.

\begin{table*}[t]
  \centering
  \caption{Ablation study where the proposed TrajFusionNet+ architecture is compared with various architectural modifications.}
  \resizebox{0.85\textwidth}{!}{%
  \begin{tabular}{cllcccccccccc}
    \toprule
    \multicolumn{2}{c}{\multirow{2}{*}{Scenario}} & \multirow{2}{*}{Branch} & \multicolumn{5}{c}{PIE} & \multicolumn{5}{c}{JAAD{\scriptsize all}} \\
    \cmidrule(lr){4-8} \cmidrule(lr){9-13}
    & & & Acc & AUC & F1 & P & R & Acc & AUC & F1 & P & R \\
    \toprule
    base & \makecell[l]{TrajFusionNet+ architecture} & \multicolumn{1}{c}{---} & 0.93 & 0.91 & 0.87 & 0.86 & 0.88 & 0.89 & 0.86 & 0.73 & 0.65 & 0.82 \\ \hline
    1 & \makecell[l]{Remove GAM branch} & GAM & 0.92 & 0.90 & 0.86 & 0.87 & 0.85 & 0.89 & 0.85 & 0.72 & 0.65 & 0.80 \\ \hline
    2 & \makecell[l]{Replace TokenGT TF with \\ GCN in GAM branch} & GAM & 0.92 & 0.91 & 0.86 & 0.83 & 0.89 & 0.89 & 0.85 & 0.72 & 0.65 & 0.80 \\ \hline
    3 & \makecell[l]{Replace SegFormer with \\ DeepLabV3 for semantic \\ segmentation} & GAM & 0.92 & 0.90 & 0.86 & 0.86 & 0.86 & 0.89 & 0.86 & 0.72 & 0.65 & 0.81 \\ \hline
    4 & \makecell[l]{Use a single VAN network \\ in VAM branch instead of \\ applying VANs sequentially } & VAM & 0.92 & 0.91 & 0.86 & 0.84 & 0.89 & 0.89 & 0.84 & 0.70 & 0.65 & 0.76 \\ \hline
    5 & \makecell[l]{Remove bounding boxes \\ from scene images} & VAM & 0.92 & 0.91 & 0.87 & 0.85 & 0.89 & 0.87 & 0.80 & 0.64 & 0.60 & 0.69 \\ \hline
    6 & \makecell[l]{Remove trajectory prediction \\ from SAM branch} & SAM & 0.91 & 0.89 & 0.84 & 0.83 & 0.85 & 0.89 & 0.86 & 0.72 & 0.64 & 0.81 \\ \hline
    7 & \makecell[l]{Remove vehicle speed \\ from input modalities} & SAM & 0.90 & 0.86 & 0.81 & 0.87 & 0.76 & 0.88 & 0.86 & 0.71 & 0.61 & 0.84 \\
    \bottomrule  
  \end{tabular}
  }
  \label{tab:ablation_study}
\end{table*}

\section*{Conclusion}

In this work, we introduced TrajFusionNet+, a novel model for pedestrian crossing intention prediction that extends our earlier TrajFusionNet architecture. The model integrates three branches: a Sequence Attention Module for pedestrian trajectory modeling; a Visual Attention Module that utilizes a visual representation of the pedestrian trajectories by overlaying observed and predicted bounding boxes onto scene images; and a new Graph Attention Module that uses a TokenGT graph transformer to capture the relational dependencies between pedestrians and traffic elements. TrajFusionNet+ achieves improved state-of-the-art performance on the PIE and JAAD datasets. Additionally, we proposed a new evaluation protocol for pedestrian crossing intention prediction, where models are trained jointly on the PIE and JAAD datasets and evaluated separately on each. This protocol showed an improved generalization capability of TrajFusionNet+ compared to existing state-of-the-art methods.

\section*{Acknowledgments}
This research was enabled in part by support provided by the Natural Sciences and Engineering Research Council of Canada (NSERC), funding reference number RGPIN-2024-05287.

{\small
\bibliographystyle{IEEEtran}
\bibliography{references}
}

\vfill

\end{document}